\documentclass[10pt,twocolumn,letterpaper]{article}

\usepackage{iccv}
\usepackage{times}
\usepackage{epsfig}
\usepackage{graphicx}
\usepackage{amsmath}
\usepackage{amssymb}

\usepackage{algorithmic}

\usepackage{graphicx}
\usepackage{amsmath}
\usepackage{amssymb}
\usepackage{booktabs}
\usepackage{algorithm2e}
\usepackage{algorithm2e}
\usepackage{comment}
\usepackage{color,soul}
\usepackage{graphicx}
\usepackage{caption}
\usepackage{subcaption}
\usepackage{array}
\usepackage{float}
\newcolumntype{P}[1]{>{\centering\arraybackslash}p{#1}}
\newcommand{\norm}[1]{\left\lVert#1\right\rVert}
\DeclareMathOperator{\EX}{\mathbb{E}}
\SetKwInput{KwInput}{Input}                
\SetKwInput{KwOutput}{Output}  

\usepackage[pagebackref=true,breaklinks=true,letterpaper=true,colorlinks,bookmarks=false]{hyperref}
\usepackage[capitalize]{cleveref}

\iccvfinalcopy 

\def\iccvPaperID{3} 

\ificcvfinal\fi

\begin{document}

\title{Complex-valued Retrievals From Noisy Images Using Diffusion Models}

\author{Nadav Torem\thanks{Equal contribution.}\ , Roi Ronen\footnotemark[1]\ , Yoav Y. Schechner\\
Viterbi Faculty of Electrical \& Computer Eng. \\
Technion, Haifa, Israel\\
{\tt\small torem@campus.technion.ac.il}
\and
Michael Elad\\
Department of Computer Science\\
Technion, Haifa, Israel\\
}

\maketitle
\ificcvfinal\thispagestyle{empty}\fi

\begin{abstract}
{
In diverse microscopy modalities, sensors measure only real-valued  intensities. 
Additionally, the sensor readouts are affected by Poissonian-distributed photon noise.
Traditional restoration algorithms typically aim to minimize the mean squared error (MSE) between the  original and recovered images. 
This often leads to blurry outcomes with poor perceptual quality.
Recently,  deep diffusion models (DDMs) have proven to be highly capable of sampling images from the a-posteriori probability of the sought variables, resulting in visually pleasing high-quality images.
These models have mostly been suggested for real-valued images suffering from Gaussian noise. 
In this study, we  generalize  annealed Langevin Dynamics, a type of DDM, to tackle the fundamental challenges in optical imaging of complex-valued objects (and real images) affected by Poisson noise. 
We apply our algorithm to various optical scenarios, such as Fourier Ptychography, Phase Retrieval, and Poisson denoising. Our algorithm is evaluated on simulations and biological  empirical data.}
\end{abstract}

\section{Introduction}

Light is a complex-valued field. During   imaging, both the phase and intensity of the field change by the captured objects and the propagation within the optical system. 
Typically,
the examined objects are assumed to have only real values.
Recovering the phase of complex-valued objects is often crucial~\cite{chan2022holocurtains, akiyama2019first}.  However, image sensors can only measure real-valued non-negative intensities. This limitation gives rise to inverse problems such as phase-retrieval~\cite{hyder2020solving,xue2022convergence,zhang2021physics,dremeau2015reference, wu2019wish} ptychography~\cite{rodenburg2019ptychography,cha2021deepphasecut}, Fourier ptychography~\cite{shamshad2019adaptive,tian2014multiplexed,saha2022lwgnet, zheng2013wide, eckert2018efficient}, coded diffraction imaging~\cite{attal2022towards} and holography. 
A common characteristic of all these problems is the {\em nonlinear} relationship between the measurements and the complex-valued unknowns.

\begin{figure}[t]
\centering
\includegraphics[width=.95\linewidth]{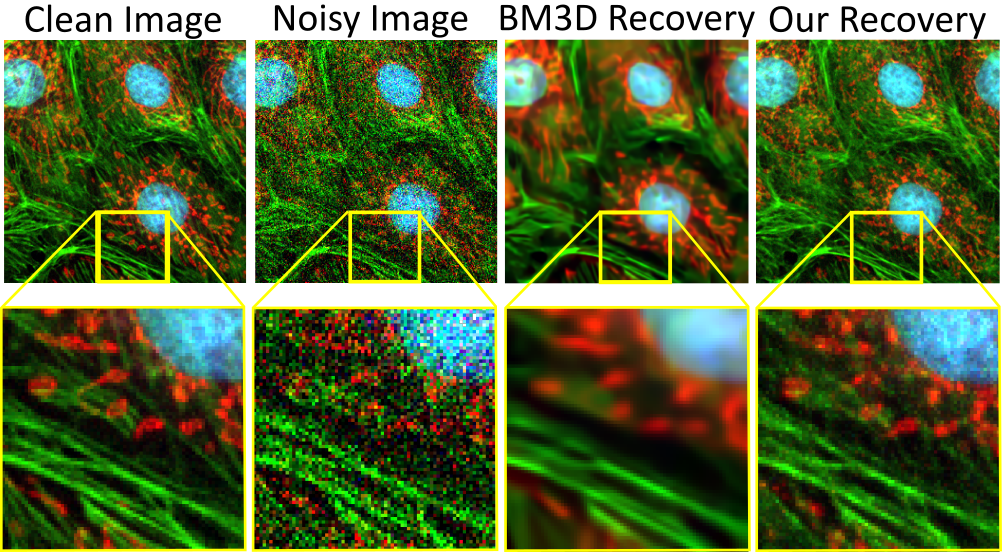}
  \caption{Poisson denoising. Results of BM3D + I + VST~\cite{azzari2016variance} and our algorithm on {real data}~\cite{zhang2019poisson}, obtained via two-photon microscopy. [Bottom]  A zoom-in.
  Our recovery has high-perceptual quality (sharp) and low distortion.
  }
  \vspace{-0.3cm}
  \label{fig:reb}
\end{figure}

Furthermore, the particle nature of light leads to Poissonian-distributed photon noise, in contrast to the commonly assumed Gaussian noise~\cite{huang2021neighbor2neighbor,kumar2019low,zhang2019poisson,byun2021fbi,moseley2021extreme,khademi2021self, mildenhall2018burst}.
Poisson noise  complicates the recovery of phase-related problems as well as linear problems involving real-valued images. 
Therefore, addressing Poisson noise during restoration is of great importance, especially in scientific tasks, such as cell imaging (see \cref{fig:reb}).
\begin{figure*}
  \includegraphics[width=\linewidth]{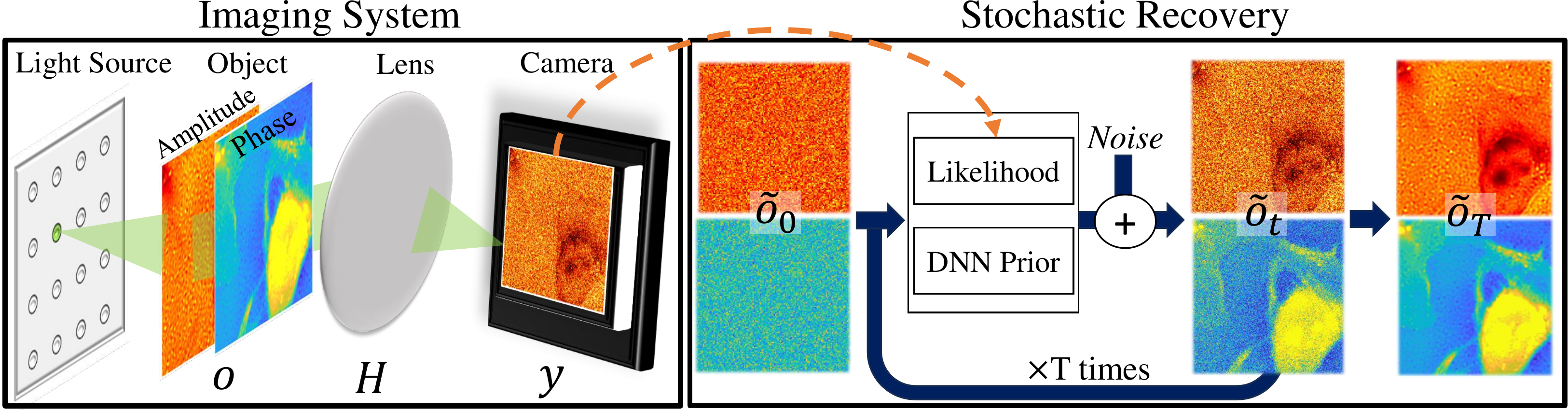}
    \vspace{-.6cm}
  \caption{Imaging of a complex-valued objects. The imaging setup captures only real valued noisy intensities. Given the noisy measurements, our method stochastically recovers 
  the objects amplitude and phase.
  } 
  \label{fig:setup}
  \vspace{-.3cm}
\end{figure*}


Over the years,  restoration algorithms have varied based on their objective. These algorithms, whether classical (NLM~\cite{buades2005non}, KSVD~\cite{elad2006image}, EPLL~\cite{zoran2011learning}, WNNM~\cite{gu2014weighted}, and BM3D~\cite{liu2018non}) or based on deep-learning (TNRD~\cite{chen2016trainable}, MALA~\cite{funke2018large}, DnCNN~\cite{zhang2017beyond}, Noise2Void~\cite{krull2019noise2void} and NLRN~\cite{liu2018non}), have been typically designed to minimize the mean squared error (MSE) between  true and estimated images. These are MMSE estimators.  
Alternatively, other methods pursued maximum a-posteriori probability (MAP)\footnote{Interestingly, also works on these methods tend to report performance using MSE.}.

The distortion-perception trade-off~\cite{blau2018perception,wang2009mean} states that in image recovery, the resulting distortion (MSE score) is reciprocal to the resulting perceptual quality. 
Hence, there is a growing interest in recovery algorithms that yield high perceptual quality, rather than minimizing the MSE. Some prior works along this line use generative adversarial networks (GANs) for image recovery. GANs can be turned into samplers from the posterior distribution.
However, training difficulties and lack of solid mathematical guarantees hindered these techniques.

Newcomer generative models are deep diffusion models (DDMs). DDMs are  a highly-performing class of iterative probabilistic generative models. 
DMMs have shown success in solving various inverse problems~\cite{song2021solving,choi2021ilvr,kawar2022denoising,chung2022improving,kadkhodaie2021stochastic,kawar2021snips,chung2022come}. However, a DMM requires a likelihood term, which is often analytically intractable. This is exacerbated by the iterative nature of DDMs.
As a result, restoration algorithms that utilize DDMs have primarily concentrated on linear inverse problems affected by stationary Gaussian noise~\cite{kawar2022denoising,kawar2021snips,kadkhodaie2021stochastic}.
An exception is~\cite{chung2022diffusion} which proposes DDMs for {\em nonlinear} inverse problems. It has two limitations: (i) Being restricted to real-valued images, and (ii) Requiring costly back-propagation through a DNN in each iteration. This  lengthens computation time during recovery.

One type of DDM is {\em annealed Langevin Dynamics}. In this study, we generalize this DDM type, to tackle fundamental challenges in optical imaging: recovering  complex-valued objects (and real images) affected by Poisson noise (see \cref{fig:setup}). 
To handle the untraceable likelihood term, we analytically develop a novel relaxed model, derived from the optical setting and the Poisson noise distribution. 
We show that the gradient of the prior can be expressed via an expectation, which is conveniently approximated by a deep neural network (DNN). 
We apply our algorithm to various optical scenarios, such as Fourier ptychography, phase retrieval, and Poisson denoising. 
Our method samples from the posterior distribution, thus exposing uncertainties in the recovery,  via the spread of the results. The uncertainty is valuable when imaging true objects.
We show results on simulations and real empirical biological data.





\section{Theoretical Background}
\subsection{Noise}
\label{sec:poisson}
Empirical light measurements are random, mainly due to the
integer nature of photons and electric charges. This randomness is  often regarded as photon noise. Dark noise, quantization noise, and photon noise are the dominant noise sources in common cameras~\cite{ratner2007illumination}. 
Denote incorporation of noise into the {expected noiseless signal} $\texttt I$ by the operator $\cal N$.
The noisy measurement is $y = {\cal N}({\texttt I})$.

Photon noise is modeled by a Poisson process~\cite{ratner2007illumination}.
A pixel is characterized by its full well capacity (FWC), that is, the maximal amount of charge that can be accumulated in the pixel before saturating. 
Let $x^{\rm e} \in  [0, {\rm FWC}]$ be the {expected (noiseless) value of a} measured signal (photo-electrons). 
The corresponding noisy measurement is
\begin{equation}
    y^{\rm e} \sim {\rm Poisson}(x^{\rm e}).
    \label{eq:poisson}
\end{equation}
Photon noise is fundamental. Its crucial nature of signal-dependency~(\ref{eq:poisson})
is valid both for photon-hungry scenes, and even more so for
well-lit scenes, as the noise variance is higher in brighter
pixels.

Modeling the nature of noise, being signal-dependent, is important when
solving scientific inverse problems in optical imaging. However, data analysis under Poisson noise is difficult to deal with in a closed form.  Therefore, in many inverse problems,  Poisson noise is approximated by a Gaussian distribution,  $n^{\rm e} \sim N(0, x^{\rm e})$. The variance still depends on the {noiseless signal}. 
Define  normalized variables 
\begin{equation}
    x=\frac{x^e}{{\rm FWC}}\in  [0, 1] \;,~~ y=\frac{y^{\rm e}}{\rm FWC}\;,~~  \sigma_0 = \frac{1}{\sqrt{{\rm FWC}}} \;.
    \label{eq:fwc}
\end{equation}
The Gaussian approximation of a {measurement} is
\begin{equation}
n^{\rm meas} \sim N(0, \sigma_0^2x)    \;.
\label{eq:n_meas}
\end{equation}
The normalized approximation of Eq.~(\ref{eq:poisson}) is 
\begin{equation}
    y \approx x + n^{\rm meas}  \;.
    \label{eq:normal_noise_model}
\end{equation}
In Eqs.~(\ref{eq:poisson},\ref{eq:normal_noise_model}), the noise variance {is linear in} $x$, and the signal-to-noise-ratio (SNR) is the same. Hence, in this work, the parameter $\sigma_0$ controls  the measurement noise level.

\subsection{Imaging of Complex-valued Objects}
\label{sec:nonlinear_imaging}
Let $\mathbf{r}$  represent the 2D spatial coordinate of an object. Let $a(\mathbf{r})$, ${\phi(\mathbf{r})}$ be its amplitude and phase, respectively, and  $j=\sqrt{-1}$. An optically-thin object has transmittance ${o}(\mathbf{r}) = {a(\mathbf{r})}\exp\left[{j {\phi(\mathbf{r})}}\right]$.
Cameras only measure intensity. Therefore, optical imaging systems can often be formulated by an imaging model, which is nonlinear in $o$. Let $\rho$ be the object irradiance. Let $H$ be a linear transformation, which  depends on the optical system. 
A noisy measurement $y$ is modeled by
\begin{equation}
y = {\cal N}\left(\rho|{H}\{o(\mathbf{r})\}|^2\right)\;.
\label{eq:non_linear_transorm}
\end{equation}
For example, in the phase retrieval problem~\cite{shechtman2015phase}, ${H}$ is equivalent to the Fourier transform ${\mathcal{F}}$.

Fourier ptychography uses a set of $M$  linear transformations ${H}=\{{H}_m\}_{m=1}^{M}$. This method enables to expand the diffraction limit set by the numerical aperture of the optical system, thus increasing the optical resolution. Let  ${\mathcal{F}}^{-1}$ 
be the inverse Fourier transform. In the spatial frequency domain $\mathbf{k}$, the pupil function ${P}(\mathbf{k}-\mathbf{k}_m)$ is a 2D bandpass filter, centered at spatial frequency $\mathbf{k}_m$ with a spatial bandwidth set by the system numerical aperture. 
Overall,
\begin{equation}
{H}_m\{o(\mathbf{r})\} = {\mathcal{F}}^{-1}\{{P}(\mathbf{k}-\mathbf{k}_m) {\mathcal{F}}\{o(\mathbf{r})\}\}.
\label{eq:ptychography}
\end{equation}

Discretize  $o(\mathbf{r})$ to $q$ elements and the measurements $y(\mathbf{r})$  to $q^{\rm meas}=Mq$ sampled pixels, creating vectors $\mathbf{o} \in \mathbb{C}^q$ and $ \mathbf{y}\in \mathbb{R}^{q^{\rm meas}}$. The discrete version of the linear transformation $H$ is $\mathbf{H} \in \mathbb{C}^{q^{\rm meas} \times q}$. The discrete versions of the Fourier transform, its inverse, and the pupil function are denoted by $ \mathbf{F},\mathbf{F}^{-1}$ and $\mathbf{P}_m$, respectively. Then, $\mathbf{H}_m\mathbf{o} = \sqrt{\rho} \mathbf{F}^{-1}\left[\mathbf{P}_m \odot\left[\mathbf{Fo}\right]\right]$,
where $\odot$ is an  element-wise multiplication. 
Assume that the dominant noise in Eq.~(\ref{eq:non_linear_transorm}) is photon noise. Following Sec.~(\ref{sec:poisson}), during data analysis, we approximate this noise as Gaussian  $\mathbf{n}^{\rm meas}\sim N(0, \sigma_0^2|\mathbf{H}\mathbf{o}|^2)$. Then, Eq.~(\ref{eq:non_linear_transorm}) takes the form 
\begin{equation}
    \mathbf{y} = |\mathbf{H}\mathbf{o}|^2 + \mathbf{n^{\rm meas}} \;.
    \label{eq:fm_non_linear}
\end{equation}
\

\subsection{Langevin Dynamics}
\label{sec:LD}
Denoising is  ill-posed: a noisy input may have multiple possible solutions. 
To address this, \emph{Langevin Dynamics} can be used for denoising, converging stochastically to a single sharp possible solution. Here we detail it shortly.

Let $t\in[1,\dots,T]$ be an iteration index and $\alpha_t>0$ be an appropriately chosen small function of $t$, having $\alpha_{t\rightarrow T}= 0$.
To maximize the probability distribution 
 $p(\mathbf{x}|\mathbf{y})$, the following iterations can be apply
\begin{equation}
\mathbf{x}_{t+1} = \mathbf{x}_t + \alpha_t\nabla_{\mathbf{x}_t}\log p(\mathbf{x}_{t}|\mathbf{y}) \;.
\label{eq:LD_nooise}
\end{equation} 
Eq.~(\ref{eq:LD_nooise})  may converge to an undesired local maximum.
To address that,  Langevin Dynamics~\cite{welling2011bayesian}  adds to Eq.~(\ref{eq:LD_nooise}) \emph{synthetic} noise \mbox{$\mathbf{n}_t^{\rm Langevin} \sim N(0,\mathbf{I}) $} at iteration $t$,  
\begin{equation}
\mathbf{x}_{t+1} = \mathbf{x}_t + \alpha_t\nabla_{\mathbf{x}_t}\log p(\mathbf{x}_{t}|\mathbf{y})+\sqrt{2\alpha_t}\mathbf{n}^{\rm Langevin}_t \;.
\label{eq:LD}
\end{equation}
Langevin Dynamics is a type of gradient descent. It relies on stochastic steps to converge to a  probable solution.

In general, when solving an inverse problem,   neither $p(\mathbf{x}_t|\mathbf{y})$ nor $\nabla_{\mathbf{x}_t}\log p(\mathbf{x}_t|\mathbf{y})$ are known.
Still, to use Eq.~(\ref{eq:LD}), there is a  need to assess $\nabla_{\mathbf{x}_t}\log p(\mathbf{x}_t|\mathbf{y})$.
To address that, Ref.~\cite{song2019generative} extends    Eq.~(\ref{eq:LD}) into \emph{annealed} Langevin Dynamics. 
Hence, iterations would \emph{not} perform  Eq.~(\ref{eq:LD}), and would not attempt to update a vector $\mathbf{x}_t$ whose probability cannot be computed.
Instead, iterations would update a vector, that we denote $\mathbf{\tilde{x}}_t$, for which we can analytically assess the gradient $\nabla_{\mathbf{\tilde{x}}_t}\log{p(\mathbf{\tilde{x}}_t|\mathbf{{y}})}$.
We thus need to properly define $\mathbf{\tilde{x}}_t$.
Analogously to Eq.~(\ref{eq:LD}),  annealed Langevin Dynamics has this iterative rule
\begin{gather}
\mathbf{\tilde{x}}_{t+1} = \mathbf{\tilde{x}}_t + \alpha_t \nabla_{\mathbf{\tilde{x}}_t}\log{p(\mathbf{\tilde{x}}_t|\mathbf{y})} +\sqrt{2\alpha_t}\mathbf{n}^{\rm Langevin}_t\;.
\label{eq:ALD_posterior}
\end{gather}
Any  $\mathbf{\tilde{x}}_t$ differs from $\mathbf{x}$ by an error $\mathbf{e}_t^{\rm annea}$,
\begin{equation}
\mathbf{\tilde{x}}_t = \mathbf{{x}} + \mathbf{e}_t^{\rm annea} \;.
\label{eq:ALD_definition}
\end{equation}
We stress that this induced error $\mathbf{e}_t^{\rm annea}$ is only a conceptual mathematical maneuver, and is not actually added during the iterative process.
It is important that the probability distribution of $\mathbf{e}_t^{\rm annea}$ would be known, despite not knowing the actual value of $\mathbf{e}_t^{\rm annea}$.
Moreover, the error \emph{is designed} such that 
$\mathbf{e}_{t\rightarrow T}^{\rm annea}\approx 0$.

Let us proceed with the description of annealed Langevin Dynamics \cref{eq:ALD_posterior}.
Using Bayes rule,
\begin{equation}
\nabla_{\mathbf{\tilde{x}}_t}\log{p(\mathbf{\tilde{x}}_t|\mathbf{y})} =\nabla_{\mathbf{\tilde{x}}_t}\left[\log{p(\mathbf{y}|\mathbf{\tilde{x}}_t)} + \log{p(\mathbf{\tilde{x}}_t)}\right] \;.
\label{eq:bayes}
\end{equation}
There is a need to derive  tractable expressions to the right-hand side of Eq.~(\ref{eq:bayes}).
The term $\nabla_{\mathbf{\tilde{x}}_t}\log{p(\mathbf{y}|\mathbf{\tilde{x}}_t)}$ in Eq.~(\ref{eq:bayes}) can sometimes  be derived analytically using the known distribution of $\mathbf{e}_t^{\rm annea}$, as we show in \cref{sec:denoising_poi,sec:recover_comlex}.
To obtain $\nabla_{\mathbf{\tilde{x}}_t}\log p(\mathbf{\tilde{x}}_t)$, Refs.~\cite{kawar2021snips,kawar2021stochastic} follow~\cite{song2020improved}:
Before imaging takes place, they pre-train a DNN denoiser
to approximate
$\nabla_{\mathbf{\tilde{x}}_t}\log p(\mathbf{\tilde{x}}_t)$.
\section{Langevin Dynamics for {Optical Imaging}}
As described in \cref{sec:poisson,sec:nonlinear_imaging}, real-world measurements have: (a) Poisson distributed noise; (b) Nonlinear transforms over complex-valued objects. There is a need to extend annealed Langevin Dynamics for both (a) and (b).

{As discussed in \cref{sec:LD}, in general inverse problems $\nabla_{\mathbf{x}}\log{p(\mathbf{x}|\mathbf{y})}$ is unknown. Therefore, in annealed Langevin Dynamics, $\nabla_{\mathbf{\tilde{x}}_t}\log{p(\mathbf{\tilde{x}}_t|\mathbf{y})}$ is exploited, while using the connection 
\begin{equation}
    \lim\limits_{t\to T} \nabla_{\mathbf{\tilde{x}}_t}\log{p(\mathbf{\tilde{x}}_t|\mathbf{y})}\rightarrow \nabla_{\mathbf{x}}\log{p(\mathbf{x}|\mathbf{y})}. 
\end{equation}
In a similar fashion, to analytically handle signal dependency of noise, we model the noise as if the signal dependency is proportional to $\mathbf{\tilde{x}}_t$ rather then $\mathbf{x}$. This allows us to analytically develop $\nabla_{\mathbf{\tilde{x}}_t}\log{p(\mathbf{y}|\mathbf{\tilde{x}}_t)}$ as described later on in \cref{sec:denoising_poi,sec:relaxation}. From \cref{eq:ALD_definition}, notice that our approximation becomes accurate as ${\tilde{\mathbf{x}}}_t\rightarrow {\mathbf{x}}$.} 
The iterative algorithm is described in \cref{alg:two}.

\RestyleAlgo{ruled}
\begin{algorithm}[t]
\caption{Langevin Dynamics for complex-valued objects. For real-valued images, $\mathbf{\tilde o}$ is replaced by $\mathbf{\tilde x}$.}
Initialize $\mathbf{\tilde o}_0$,
set a step size  $\epsilon>0$ 
\\
\For{$t \gets 1 $ to T}{
    $\alpha_t \gets \epsilon({\sigma^2_t}/{\sigma^2_{T}})$\\

        Draw $\mathbf{n}^{\rm Langevin}_t$ from a Gaussian distribution\\
        \eIf{$\mathbf{\tilde o}$ is complex-valued}
         {From  \cref{eq:complex_final} compute gradient
         \\ \mbox{$\mathbf{\Delta}_t \gets \! \nabla_{\mathbf{{\tilde o}}_{t-1}}\log p(\mathbf{y}|\mathbf{{\tilde o}}_{t-1})  +
        h_{\hat \Theta}^{\rm complex}(\mathbf{\tilde o}_{t-1}|\sigma_t)$ }} 
             {From  \cref{eq:gradient_poisson} compute gradient
             \\
             \mbox{$\mathbf{\Delta}_t \gets \! \nabla_{\mathbf{{\tilde o}}_{t-1}}\log p(\mathbf{y}|\mathbf{{\tilde o}}_{t-1})  +
        h_{\hat \Theta}^{\rm Poisson}(\mathbf{\tilde o}_{t-1}|\sigma_t)$ }}  
        $\mathbf{\tilde o}_t \gets \mathbf{\tilde o}_{t-1} +\alpha_t\mathbf{\Delta}_t + \sqrt{2\alpha_t}\mathbf{n}^{\rm Langevin}_t$
}
$\hat {\mathbf{o}} \gets \mathbf{\tilde o}_T$
\label{alg:two}
\end{algorithm}

\subsection{Denoising Poissonian Image Intensities}
\label{sec:denoising_poi}

We now attempt to recover an image $\mathbf{x}$, given its noisy version $\mathbf{y}$. As discussed in \cref{sec:poisson}, $\mathbf{y}$ contains several types of noise. Indeed data is {\emph{generated}} using a {rich noise model: true Poissonian, compounded by graylevel quantization.} For {\emph{data analysis}}, however, we  act as if $\mathbf{y}$ is related by \cref{eq:normal_noise_model}.
The parameter $\sigma_0$ is known from the camera properties by \cref{eq:fwc}, as described in \cref{sec:poisson}.  
Here, all vector operations are \emph{element-wise}.
Analogously to~\cite{kawar2021snips,kawar2021stochastic}, we set a sequence of noise levels 
{$\sigma_0 > \sigma_1 \ge \sigma_2 \ge \dots \ge \sigma_{T} > \sigma_{T+1} =  0$.}
Let  $\mathbf{e}^{\rm part}_t$ be a statistically independent  synthetic \emph{partial} dummy error term, distributed as
\begin{equation}
 \mathbf{e}^{\rm part}_t \sim N\left[0,(\sigma^2_t - \sigma^2_{t+1})\mathbf{I}\right]   \;.
\end{equation}
A sum of zero-mean Gaussian variables is a Gaussian, whose variance is the sum of the  variances. So,
\begin{equation}
 \sum^{t_2}_{\tau=t_1}\mathbf{e}^{\rm part}_\tau \sim\ N\left[0,(\sigma^2_{t_1} - \sigma^2_{t_2})\mathbf{I}\right]\;.
\label{eq:sum_part} 
\end{equation}
Using  \cref{eq:n_meas,eq:sum_part,,eq:n_meas} , we design $\mathbf{e}_t^{\rm part}$ such that
\begin{equation}
\mathbf{{{n}}}^{\rm meas} = \sqrt{\mathbf{x}}\sum^T_{\tau=0}\mathbf{e}^{\rm part}_\tau  \;.
\label{eq:w_hat}
\end{equation}
Moreover, recall that $\mathbf{e}_t^{\rm annea}$ is a dummy error term. We set  $\mathbf{e}_t^{\rm annea}$ in this task as
\begin{equation}
\mathbf{e}_t^{\rm annea} = \sum^{T}_{\tau=t}\mathbf{e}^{\rm part}_\tau \;.
\label{eq:n_eta} 
\end{equation}
We do \emph{not} set $\mathbf{e}_t^{\rm annea}$ as Refs.~\cite{kawar2021snips,kawar2021stochastic} because the noise in Refs.~\cite{kawar2021snips,kawar2021stochastic}  has stationary Gaussian distribution and here it is Poissonian (signal-dependent).


We generalize \cref{eq:ALD_definition} to Poisson noise, by defining  an annealed variable
\begin{equation}
\mathbf{\tilde{x}}_t = \mathbf{x} + \sqrt{\mathbf{x}}\mathbf{e}^{\rm annea}_t =
\mathbf{x} + \sqrt{\mathbf{x}}\sum^{T}_{\tau=t}\mathbf{e}^{\rm part}_\tau
\;.
\label{eq:defining_x_tilde}
\end{equation}
 \cref{eq:defining_x_tilde} allows us to derive $\nabla_{\mathbf{\tilde{x}}_t}\log p(\mathbf{y}|\mathbf{\tilde{x}}_t)$ analytically for Poisson denoising.
Inserting \cref{eq:defining_x_tilde,,eq:w_hat}  in \cref{eq:normal_noise_model} yields
\begin{equation}
\mathbf{y} = \mathbf{\tilde{x}}_t - \sqrt{\mathbf{x}} \sum^T_{\tau=t}\mathbf{e}^{\rm part}_\tau  +\sqrt{\mathbf{x}} \sum^T_{\tau=0}\mathbf{e}^{\rm part}_\tau 
= \mathbf{\tilde{x}}_t+ \sqrt{\mathbf{{x}}}\sum^{t-1}_{\tau=0}\mathbf{e}^{\rm part}_\tau.
\label{eq:y_x_tilde}
\end{equation}

Notice that $\mathbf{\tilde{x}}_t$ is statistically independent of the additional dummy error $\sum^{t-1}_{\tau=0}\mathbf{e}^{\rm part}_\tau$. 
We apply the following steps.
From \cref{eq:y_x_tilde}, 
\begin{gather}
            {p(\mathbf{y}|\mathbf{\tilde{x}}_t)} =  
            {p(\mathbf{y} - \mathbf{\tilde{x}}_t|\mathbf{\tilde{x}}_t)} = 
            {p \left(\left[\sqrt{\mathbf{{x}}}\sum^{t-1}_{\tau=0}\mathbf{e}^{\rm part}_\tau\right]|\mathbf{\tilde{x}}_t\right)}.
            \label{eq:gt_likelihood_poisson}
\end{gather}        
         %
{Recall that ${\mathbf{x}}$ in \cref{eq:gt_likelihood_poisson} is unknown. Therefore, to set an explicit analytical expression of \cref{eq:gt_likelihood_poisson}, we approximate the remaining measurement noise as if the signal dependency is proportional to $\mathbf{\tilde{x}}_t$ rather\footnote{From \cref{eq:defining_x_tilde}, this approximation becomes accurate as  ${\tilde{\mathbf{x}}}_t\rightarrow {\mathbf{x}}$.}} then $\mathbf{x}$.
{Overall, \cref{eq:gt_likelihood_poisson} takes the form of
\begin{equation}
    p(\mathbf{y}|\mathbf{{\tilde{x}}}_t) \approx {p \left(\left[\sqrt{{\mathbf{{\tilde{x}}}}}\sum^{t-1}_{\tau=0}\mathbf{e}^{\rm part}_\tau\right]|\mathbf{\tilde{x}}_t\right)}.
\end{equation}
Finally, the expression $\sqrt{\mathbf{{\tilde{x}}}_t}\sum^{t-1}_{\tau=0}\mathbf{e}^{\rm part}_\tau$ is both known and statistically tractable. This allows us to find an explicit expression of the probability distribution $p(\mathbf{y}|\mathbf{\tilde{x}}_t)$
\begin{equation}
           \!\!\!\! {{ {p}}(\mathbf{y}|\mathbf{\tilde{x}}_t)}  
         \! \approx\! \frac{1}{\sqrt{2\pi(\sigma{^2_0} - \sigma{^2_t})\mathbf{\tilde{x}}}_t}\exp\left[{\!-\frac{({\mathbf{y} - \mathbf{\tilde{x}}_t})^2}{2(\sigma{^2_0} - \sigma{^2_t}){\mathbf{\tilde{x}}}_t}}\right].
         \label{eq:likelihood_poisson}
\end{equation}}
Computing the gradient of the logarithm of \cref{eq:likelihood_poisson} yields
\begin{gather}
 \nabla_{\mathbf{\tilde{x}}_t}\log { p}(\mathbf{y}|\mathbf{\tilde{x}}_t)=  
    \frac{1}{2(\sigma{^2_0} - \sigma{^2_t})}
    \left(\frac{\mathbf{y}^2} {\mathbf{\tilde{x}}^2_t} - 1\right) -\frac{1}{2\mathbf{\tilde{x}}_t} \;.
    \label{eq:log_p_y_x_tilde}
\end{gather}

In analogy to \cref{sec:LD}, we compute $\nabla_{\mathbf{\tilde{x}}_t}\log{p(\mathbf{\tilde{x}}_t)}$.
The derivation  is detailed in the supplementary material.
Here we bring the result: The gradient $\nabla_{\mathbf{{\tilde{x}}}_t}\log p(\mathbf{{\tilde{x}}}_t)$ is given by,  
\begin{equation}
    \nabla_{\mathbf{{\tilde{x}}}_t}\log p(\mathbf{{\tilde{x}}}_t) = \EX\left[{\frac{\mathbf{{x}}-\mathbf{{\tilde{x}}}_t}{{\sigma^2}\mathbf{{x}}}}|\mathbf{\tilde{x}}_t\right].
    \label{eq:true_p}
\end{equation}
The expectation in~\cref{eq:true_p} cannot be calculated analytically. The reason is that the probability distribution of $\mathbf{{x}}$ is not accessible.  So, before imaging takes place, we pre-train a DNN  $h_\Theta^{\rm Poisson}$ having learnable parameters $\Theta$ to estimate $\nabla_{\mathbf{{\tilde{x}}}_t}\log p(\mathbf{{\tilde{x}}}_t)$, using synthetic pairs of corresponding clean and noisy images $(\mathbf{x},\mathbf{x}')$,
\begin{equation}
\hat {\Theta} = \arg \min_\Theta
\EX\left[\norm{\frac{\mathbf{{x}}-\mathbf{{x}}'}{{\sigma^2}\mathbf{{x}}} - h_\Theta^{\rm Poisson}(\mathbf{{x}'}|\sigma)}^2_2\right] \;. 
\label{eq:DNN_poisson}
\end{equation}
The expectation $\EX$ is approximated by an empirical average of all pairs in the training set of corresponding clean and noisy images $(\mathbf{x},\mathbf{x}')$.
Overall, Eqs.~(\ref{eq:log_p_y_x_tilde},\ref{eq:true_p},\ref{eq:DNN_poisson}) yeild
\begin{gather}
\nabla_{\mathbf{\tilde{x}}_t}\log p(\mathbf{\tilde{x}}_t|\mathbf{y}) = \nabla_{\mathbf{\tilde{x}}_t}\log p(\mathbf{\mathbf{y}|\tilde{x}}_t) + h_{\hat \Theta}^{\rm Poisson}(\mathbf{\tilde{x}}_t|\sigma_t) = \nonumber\\ \frac{1}{2(\sigma{^2_0} - \sigma{^2_t})}
    \left(\frac{\mathbf{y}^2} {\mathbf{\tilde{x}}^2_t} - 1\right) -\frac{1}{2\mathbf{\tilde{x}}_t} +h_{\hat \Theta}^{\rm Poisson}(\mathbf{\tilde{x}}_t|\sigma_t) \;.
    \label{eq:gradient_poisson}
\end{gather}
\subsection{Recovering Complex-Valued Objects}
\label{sec:recover_comlex}

This section formulates annealed 
Langevin Dynamics for {complex-valued,} nonlinear imaging models, as described in \cref{sec:nonlinear_imaging}. 
Let ${\Re}(\cdot)$, $\Im(\cdot)$ be real and imaginary arguments of a complex-valued number, respectively.
Analogously to \cref{eq:LD}, for  Langevin Dynamics,  $\nabla_{\mathbf{{o}}_t}\log{p(\mathbf{{o}}_t|\mathbf{y})}$ is needed. However, $p(\mathbf{{o}}_t|\mathbf{y})$ is not known, thus analogously to \cref{eq:ALD_posterior}, annealed Langevin Dynamics for a complex-valued object is
\begin{gather}
\mathbf{\tilde{o}}_{t+1} = \mathbf{\tilde{o}}_t + \alpha_t \nabla_{\mathbf{\tilde{o}}_t}\log{p(\mathbf{\tilde{o}}_t|\mathbf{y})} +\sqrt{2\alpha_t}\mathbf{n}^{\rm Langevin}_t\;.
\label{eq:ALD_posterior_complex}
\end{gather}
Here, $\mathbf{n}^{\rm Langevin}$ is a random complex-valued vector, where $\Re(\mathbf{n}^{\rm Langevin}),\Im(\mathbf{n}^{\rm Langevin})\sim N(0,\mathbf{I})$ and independent.
For simplicity, we present  here the special case  $\mathbf{H = I}$ in \cref{eq:fm_non_linear}. Thus, most derivations can be given in scalar formulation,  that is, ${{y}} = |{{o}}|^2 + {{{n}^{\rm meas}}}$. The generalization of the model to a general $\mathbf{H}$ is detailed in the supplementary material.
Analogously to \cref{eq:ALD_definition}, we set
\begin{equation}
    \tilde{{o}}_t = {{o}} + {e}_t^{\rm annea} \;.
    \label{eq:o_t}
\end{equation}   
Using \cref{eq:o_t} in \cref{eq:fm_non_linear} yields 
\begin{gather}
{{y}} = |\tilde{o}_t|^2 
- {e}_t^{\rm couple} + |{e}^{\rm annea}_t|^2+ {{{n}^{\rm meas}}} \;,
\label{eq:y_o_long}
\end{gather}   
where 
\begin{gather}
    |{o}_t|^2=\Re({o}_t)^2+\Im({o}_t)^2\;,
    \label{eq:abs_o}
\end{gather}
\begin{equation}
    {e}_t^{\rm couple} =  2\left[\Re({\tilde{o}}_t)  \Re({e}_t^{\rm annea})+ \Im({\tilde{o}}_t)  \Im({e}_t^{\rm annea})\right] \;,
    \label{eq:e_couple}
\end{equation}
and $|{e}_t^{\rm annea}|^2=\Re({e}_t^{\rm annea})^2+\Im({e}_t^{\rm annea})^2$.
The non-linearity of \cref{eq:fm_non_linear} yields two  terms in \cref{eq:y_o_long} that pose  analytic challenges:  a coupling term of ${e}_t^{\rm couple}$;
and a squared annealing error term $|{e}^{\rm annea}_t|^2$. 
We handle them by introducing a \emph{relaxed}  model, creating annealed  Langevin Dynamics for complex-valued objects,  as
\begin{equation}
{{y}} =  |\tilde{o}_t|^2 
- {e}_t^{\rm couple} + |{e}^{\rm annea}_t|^2 + {\tilde{n}}_t^{\rm meas} -\sigma^2_t{z}_t \;.
\label{eq:relaxed_nonlinear_LD}
\end{equation}
We detail \cref{eq:relaxed_nonlinear_LD} and specifically ${{e}}_t^{\rm annea},{\tilde{n}}_t^{\rm meas}$ and ${z}_t$ in \cref{sec:relaxation}.
\begin{table*}[t]
\centering
\setlength{\textfloatsep}{0pt plus 0pt minus 0pt}

\resizebox{1\textwidth}{!}{\begin{tabular}
{P{85pt}|P{36pt}P{39pt}P{17pt}|P{36pt}P{39pt}P{17pt}|P{36pt}P{39pt}P{17pt}|P{36pt}P{39pt}P{17pt}}
 \multicolumn{1}{c|}{} 
 &\multicolumn{3}{c|}{$\sigma_0=0.025$}
 &\multicolumn{3}{c|}{$\sigma_0=0.05$}
 &\multicolumn{3}{c|}{$\sigma_0=0.1$}
 &\multicolumn{3}{c}{$\sigma_0=0.2$}
 \\
        CelebA 64x64 &PSNR$\uparrow$        & SSIM\% $\uparrow$       & FID$\downarrow$          & PSNR$\uparrow$        & SSIM\% $\uparrow$       & FID$\downarrow$ 
         & PSNR$\uparrow$        & SSIM\% $\uparrow$       & FID$\downarrow$ 
          & PSNR$\uparrow$        & SSIM\% $\uparrow$       & FID$\downarrow$ 
          \\ \hline
\!\!\!I + VST + BM3D~\cite{azzari2016variance}
&  ${39.1\!\pm\!1.0}$          &  ${99\!\pm\!0.1}$          & {2.80} 
&  ${35.8\!\pm\!1.1}$          &  ${98\!\pm\!0.2}$   & {8.90}
&  ${30.9\!\pm\!1.2}$          &  ${96\!\pm\!1.0}$   & {25.4}
&  ${27.2\!\pm\!1.1}$          &  ${93\!\pm\!2.0}$          & {63.2} 
\\
PnP RED \cite{romano2017little}
&  ${39.8\!\pm\!0.9}$          &  ${99\!\pm\!0.1}$          & {4.00} 
&  ${36.9\!\pm\!1.0}$          &  ${98\!\pm\!0.3}$   & {7.50}
&  ${31.8\!\pm\!0.9}$          &  ${96\!\pm\!1.0}$   & {12.4}
&  ${27.8\!\pm\!1.2}$          &  ${93\!\pm\!2.0}$          & {28.4} 
\\
Bahjat et al.~\cite{kawar2021stochastic}
  & $34.2\!\pm\!1.1$           &  $97\!\pm\!0.9$          &    2.10
 & $34.1\!\pm\!0.7$           &  $97\!\pm\!1.0$          &     6.70
 & $30.6\!\pm\!0.9$           &  $95\!\pm\!2.0$          &     11.1
  & $27.1\!\pm\!1.2$           &  $91\!\pm\!4.0$          &    15.2
  \\
MMSE
&  ${42.5\!\pm\!1.0}$          &  ${99\!\pm\!0.1}$          & \underline{1.32} 
&  ${38.4\!\pm\!1.1}$          &  ${99\!\pm\!0.2}$   & \underline{3.60}
&  ${34.5\!\pm\!1.1}$          &  ${98\!\pm\!0.5}$   & \underline{7.50}
&  ${30.9\!\pm\!1.2}$          &  ${96\!\pm\!1.0}$          & \textbf{11.9} 
\\
Ours    
&   ${41.5\!\pm\!1.1}$         & ${99\!\pm\!0.0}$           &  \textbf{1.25}    
&   ${37.1\!\pm\!1.0}$        & ${99\!\pm\!0.4}$           &  \textbf{2.70}    
&   ${32.8\!\pm\!1.1}$        & ${98\!\pm\!0.5}$           &  \textbf{6.10}    
&   ${28.6\!\pm\!1.3}$         & ${94\!\pm\!1.0}$           &  \underline{13.0}    
        \\ \hline
 FFHQ 256x256   
        \\ \hline
DPS~\cite{chung2022diffusion}
&  ${36.8\!\pm\!4.7}$          &  ${99\!\pm\!0.1}$          & \underline{2.00} 
&  ${33.8\!\pm\!3.8}$          &  ${98\!\pm\!0.4}$   & \underline{5.47}
&  ${31.5\!\pm\!2.7}$          &  ${98\!\pm\!2.0}$   & \underline{12.3}
&  ${28.6\!\pm\!2.0}$          &  ${93\!\pm\!2.0}$          & \textbf{23.8} 
\\
Ours    
&   ${41.9\!\pm\!1.0}$         & ${99\!\pm\!0.0}$           &  \textbf{0.72}    
&   ${37.9\!\pm\!1.3}$        & ${99\!\pm\!0.1}$           &  \textbf{2.50}    
&   ${33.5\!\pm\!1.1}$        & ${96\!\pm\!1.0}$           &  \textbf{10.0}    
&   ${29.0\!\pm\!1.0}$         & ${93\!\pm\!1.0}$           &  \underline{33.6}    
\end{tabular}}

  
     \caption{Results on 1000 images of CelebA and FFHQ datasets. {
     For $\sigma_0 \le 0.1$, our algorithm has better FID and similar SSIM with the MMSE denoiser and DPS~\cite{chung2022diffusion}. The MMSE denoiser inherently achieves superior PSNR.}}
    \label{tab:PSNRTables} 
\end{table*}
\subsection{Relaxation of the Nonlinear Problem}
\label{sec:relaxation}
Let $\{{\sigma}_t\}^{T}_{t=0}$ be a sequence of standard deviations, set to {$\sigma_0 > \sigma_1 \ge \sigma_2 \ge \dots \ge \sigma_{T} > \sigma_{T+1}=  0$.}
Let 
\begin{equation}
{\Re}({e}^{\rm part}_t), \Im({e}^{\rm part}_t) \sim N[0,\frac{1}{4}(\sigma^2_t - \sigma^2_{t+1})]   
\label{eq:part_complex}
\end{equation}
 be statistically independent  Gaussian variables. 
As described in \cref{sec:nonlinear_imaging} and in \cref{eq:n_meas}, for $\mathbf{H = I}$,  \mbox{${n}^{\rm meas}\sim N(0, \sigma_0^2|{o}|^2)$}. 
Using $|o|^2$ from \cref{eq:abs_o},
\mbox{${n}^{\rm meas}\sim N[0, \sigma_0^2\Re({o})^2+\sigma_0^2\Im({o})^2]$}. 
In analogy to \cref{eq:w_hat}, we use \cref{eq:part_complex} and design  dummy variables  ${\Re}({e}^{\rm part}_t), \Im({e}^{\rm part}_t)$ such that,
\begin{equation}
{{n}^{\rm meas}} = 2\Re({{o}})\sum^{T}_{\tau=0}{\Re}({e}^{\rm part}_\tau) + 2\Im({{o}})\sum^{T}_{\tau=0}\Im({e}^{\rm part}_\tau) \;.
\label{eq:w_pty}
\end{equation}
Notice that \cref{eq:w_pty} now has a form similar to the coupling term ${e}_t^{\rm couple}$ in \cref{eq:e_couple}.
To make computations tractable, we assume $ {o} \approx {\tilde{o}}_t$ in \cref{eq:w_pty}.  We do not use ${\tilde{o}}_t$ as a denoised version yet. An approximate\footnote{From \cref{eq:o_t}, the approximation becomes accurate as ${e}_t^{\rm annea} \rightarrow 0$, where ${\tilde{o}}_t\rightarrow {o}$ and  ${{\tilde{n}^{\rm meas}}}_t \rightarrow {n^{\rm meas}}$.}  measurement noise is
\begin{equation}
{\tilde{n}^{\rm meas}}_t = 2\Re(\tilde{{{o}}}_t)\sum^{T}_{\tau=0}{\Re}({e}^{\rm part}_\tau) + 2\Im(\tilde{{{o}}}_t)\sum^{T}_{i=0}{\Im}({e}^{\rm part}_\tau) \;.
\label{eq:w_hat_pty}
\end{equation}
Using \cref{eq:e_couple,eq:w_hat_pty}, define
\begin{gather}
{e}_t^{\rm remain}(\tilde{{{o}}}_t)= {\tilde{n}^{\rm meas}}_t - {e}_t^{\rm couple} ~~~~~~~~~~~~~~~~~~~~~~~~~~~~~~~~~~~~~~~~~~~~~~~~~~~~~~ \nonumber\\
=2\Re(\tilde{{{o}}}_t)\sum^{t-1}_{\tau=0}\Re{({e}^{\rm part}_{\tau})} + 2\Im(\tilde{{{o}}}_t)\sum^{t-1}_{\tau=0}\Im{({e}^{\rm part}_{\tau})}
\;.
\label{eq:n_couple}
\end{gather}

The term ${z}_t$ in \cref{eq:relaxed_nonlinear_LD} is designed to address the challenge of a squared annealing error term $|{e}^{\rm annea}_t|^2$. We detail  ${z}_t$ in the supplementary material,
here we give the final result,
\begin{gather}
{z}_t^{\rm remain} =   |{e}^{\rm annea}_t|^2 -\sigma_t^2{z}_t ~~~~~~~~~~~~~~~~~~~~~~~~~~~~~~~~~~~~~~~~~~~~~~~~~~~~~~~~~~~~~~~  \nonumber\\
={\left[\sum^{t-1}_{\tau=0}\Re{({e}^{\rm part}_{\tau})}\right]^2}
   \!\!\! + {\left[\sum^{t-1}_{\tau=0}\Im{({e}^{\rm part}_{\tau})}\right]^2}\;.
\label{eq:z_remain}
\end{gather}
Insert  \cref{eq:n_couple,eq:z_remain} into the relaxed model in \cref{eq:relaxed_nonlinear_LD}. This yields
\begin{equation}
{{y}} = |{\tilde{o}}_t|^2 + {e}_t^{\rm remain}(\tilde{{{o}}}_t)
+ {z}_t^{\rm remain}
\;.
\label{eq:pty_assumption}
\end{equation}
Using \cref{eq:pty_assumption},  $\nabla_{{\tilde{o}}_t}\log p({y}|{\tilde{o}}_{t})$ can be derived analytically.
Here we give the results: the derivation is detailed in the supplementary material.
Let $I_0, I_1$ be the modified Bessel functions of the first kind~\cite{bowman2012introduction}.
Then,
\begin{equation}
    \nabla_{{\tilde{o}}_t}\log p({y}|{\tilde{o}}_{t}) =  {\frac{{\tilde{o}}_{t}}{2(\sigma^2_0 - \sigma^2_t)} \left[ {\cal I}({\tilde{o}}_t,{y})\frac{\sqrt{{y}}}{ |{\tilde{o}}_{t}|} - 1\right]}
\;,
\label{eq:complex_final}
\end{equation} 
where ${\cal I}({\tilde{o}}_t,{y})={I_1\left({\frac{|{\tilde{o}}_{t}|\sqrt{{y}}}{\sigma^2_0 - \sigma^2_t}}\right)}/{I_0{\left({\frac{|{\tilde{o}}_{t}|\sqrt{{y}}}{\sigma^2_0 - \sigma^2_t}}\right)}}$.
The  gradient $ \nabla_{\mathbf{{\tilde{o}}}_t}\log p(\mathbf{{\tilde{o}}}_t)$ is given by
\begin{equation}
     \nabla_{\mathbf{{\tilde{o}}}_t}\log p(\mathbf{{\tilde{o}}}_t) =  \EX\left[\frac{\mathbf{o}-\mathbf{{\tilde{o}}}_t}{\sigma^2}|\mathbf{\tilde{o}}_t\right] \;.
     \label{eq:grad_o}
\end{equation}
See the supplemental material for the proof of \cref{eq:grad_o}.

Hence again, we pre-train\footnote{The network input is a two-channel tensor of $\Re(\mathbf{{o}'}),\Im(\mathbf{{o}'})$.} a DNN $h_\Theta^{\rm complex}$ having learnable parameters $\Theta$  to estimate  $ \nabla_{\mathbf{{\tilde{o}}}_t}\log p(\mathbf{{\tilde{o}}}_t)$
\begin{equation}
\hat {\Theta} = \arg \min_\Theta
\EX\left[\norm{\frac{\mathbf{{o}}-\mathbf{{o}}'}{{\sigma^2}} - h_\Theta^{\rm complex}(\mathbf{{o}}'|\sigma)}^2_2\right] \;. 
\label{eq:denoiser_complex}
\end{equation}
Overall, the complex-valued derivative is
\begin{gather}
\nabla_{\mathbf{\tilde{o}}_t}\log p(\mathbf{\tilde{o}}_t|\mathbf{y}) = \nabla_{\mathbf{\tilde{o}}_t}\log p(\mathbf{\mathbf{y}|\tilde{o}}_t) + h_{\hat \Theta}^{\rm complex}(\mathbf{\tilde{o}}_t|\sigma_t) =  \nonumber\\ {\frac{{\tilde{\mathbf o}}_{t}}{2(\sigma^2_0 - \sigma^2_t)}\odot \left[ {\cal I}({\tilde{\mathbf o}}_t,{\mathbf y})\odot\frac{\sqrt{{\mathbf y}}}{ |{\tilde{\mathbf o}}_{t}|} - 1\right]} +h_{\hat \Theta}^{\rm complex}(\mathbf{\tilde{o}}_t|\sigma_t) \;.
    \label{eq:gradient_complex}
\end{gather}

\section{Experiments}
\label{sec:examples}
To evaluate recoveries we use the FID, SSIM and PSNR\footnote{For complex-valued objects, the amplitude  is evaluated by standard PSNR metric. The recovered object's phase ${\angle}\mathbf{o}$  has an inherent  ambiguity due to the $|\cdot|^2$ operation and the periodicity of $2\pi$. 
Therefore, we evaluate phase recovery by
\mbox{${\rm PSNR}^{\rm phase}({\angle}\hat{\mathbf{o}}) = 
10\log_{10}\left[{(2\pi)^2}\left[{\rm MSE}^{\rm phase}({\angle}\hat{\mathbf{o}})\right]^{-1}\right]$}, where \mbox{${\rm MSE}^{\rm phase} ({\angle}\hat{\mathbf{o}})= \arg\min_\theta \norm{{\angle}\mathbf{o} - ( {\angle}\hat{\mathbf{o}}+\theta)}^2_2 /q$.}}
criteria. 
In all experiments we used the NCSNv2 architecture~\cite{song2020improved} as the DNN. {
The bottleneck of the proposed algorithm is iteratively running this DNN  $T\sim1000$ times, where time complexity increases with image size and linearly with $T$.  
}
{We apply our algorithm on both real biological data \cite{zhang2019poisson, tian2014multiplexed} and on synthetic datasets, CelebA ($64\times 64$) and FFHQ ($256\times 256$).}
We stress that during testing on synthetic scenes, we synthesized data using
{\textbf{actual Poisson noise}} (Eq.~\ref{eq:poisson}). Moreover, 8-bit quantization was applied on the
simulated electron count.
\subsection{Denoising Poissonian Image Intensities}
\label{sec:res_poisson}

\begin{figure}[t]
\centering
   \includegraphics[width=1\linewidth]{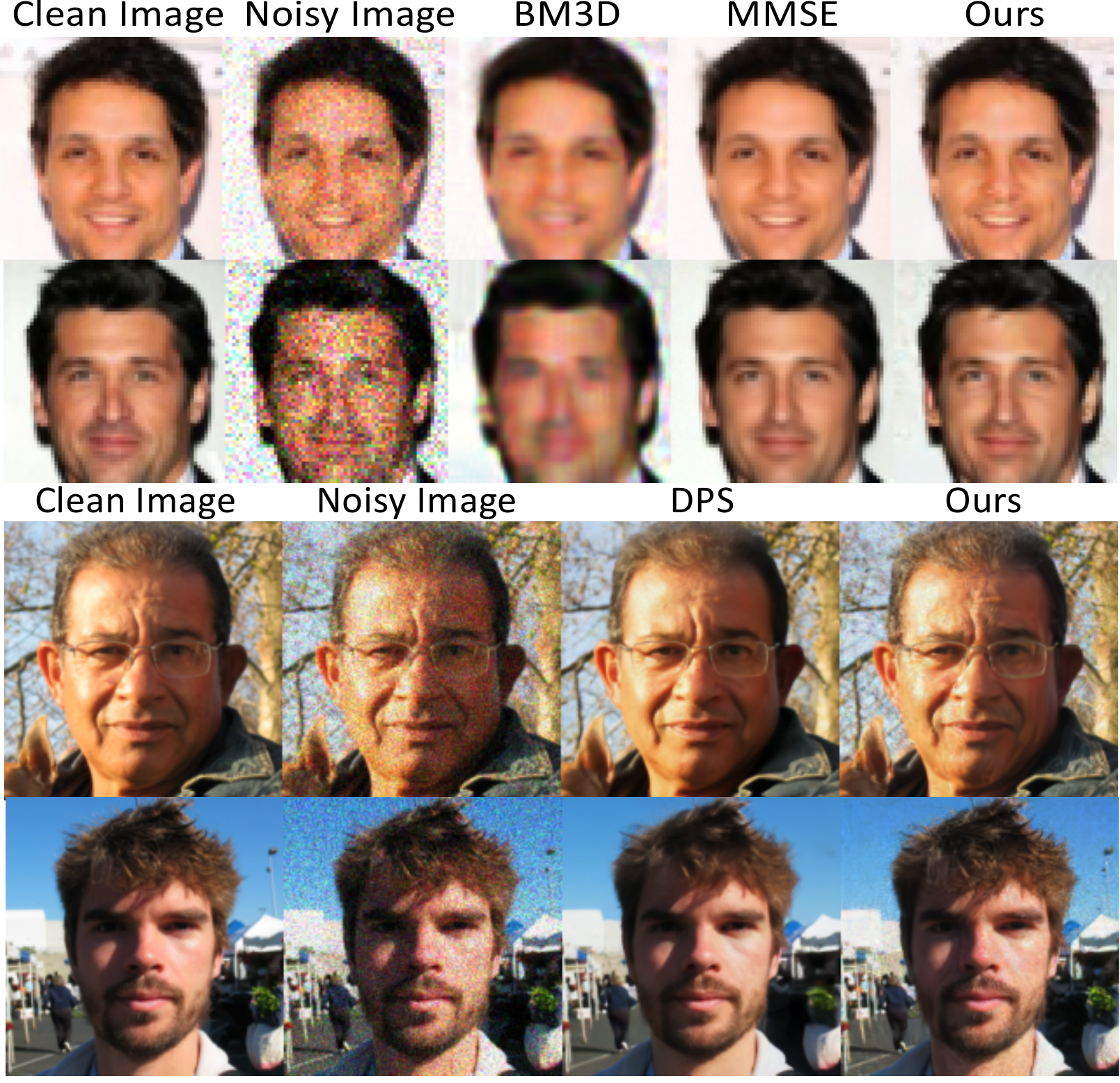}
  \caption{Poisson denoising results. From top to bottom: CelebA $\sigma_0 = 0.1$, $\sigma_0 = 0.2$. FFHQ $\sigma_0 = 0.2$. } 
  \label{fig:poisson_images} 
\end{figure}

{We compare our algorithm with  \cite{azzari2016variance,chung2022diffusion,kawar2021stochastic,romano2017little} and an MMSE solution.
\cref{fig:poisson_images} and \cref{tab:PSNRTables} present Poisson denoising results. Across all tests, our method outperforms~\cite{azzari2016variance}, a   non-learning-based state-of-the-art method for Poisson denoising. Furthermore, we outperform Ref.~\cite{kawar2021stochastic}, which utilizes annealed Langevin Dynamics for \emph{Gaussian} noise. This indicates the benefit of addressing the natural property of imaging noise. 
PnP RED~\cite{romano2017little} is a method to solve general inverse problems. It uses Gaussian denoiser as its prior\footnote{The data term derives from a Poissonian model.}. Our method surpasses it, emphasizing the significance of employing a customized denoiser for Poisson denoising.}

Typically, the SNR in standard optical systems satisfies $\sigma_0 \le 0.1$. In that regime, we have better FID results and similar SSIM results than the MMSE denoiser and DPS~\cite{chung2022diffusion}.
MMSE denoisers are designed to minimize the MSE criterion, that is,  maximizing the PSNR. Thus, unsurprisingly, an MMSE denoiser yields a superior PSNR than other algorithms, however, with blur outputs (see \cref{fig:poisson_images}).

\subsubsection{Poisson Denoising of Real Data}
\cref{fig:reb} shows  our method on real data~\cite{zhang2019poisson}. 
It contains noisy colored images of fixed BPAE cells, obtained using two-photon microscopy.
The imaged sample was split into 20 non-overlapping patches, 19 for training, and the remaining for testing.
In~\cite{zhang2019poisson}, each patch was repeatedly captured 50 times, as 50 noise realizations. 
The clean ground-truth  image is based on temporal averaging of the 50 images. 
The DNN trains only on 19  images  with  synthetic Poisson noise, while testing  on raw data involves real noise.
{The authors in Ref.~\cite{zhang2019poisson} modeled the imaging system noise by two additive components with zero means: A signal-dependent Poisson noise, and a negligible\footnote{Let $\mathbf{x}\in [0,1]$ be the true normalized image. Let $n^G$, $n^P$ be  Gaussian noise and Gaussian approximation for the Poisson noise, with corresponding standard deviations  $\sigma_{G} = 5\times10^{-4}$ and $ \sigma_{P} = \sqrt{x}(2.5\times10^{-2})$. Then, the measured image $\mathbf{y}$ is approximated by $\mathbf{y} = \mathbf{x} + n^G + n^P$.
} signal-independent Gaussian noise. Therefore, we apply our denoising algorithm as the measured image $\mathbf{y}$ can be described by \cref{eq:normal_noise_model}. }
The PSNR and SSIM of the noisy raw data  are 26 and 0.67, respectively, while   ours are  33 and 0.92.
 
\subsection{Phase Retrieval}
As described in \cref{sec:nonlinear_imaging}, phase retrieval can be modeled using $\mathbf{H} = \mathbf{F}$, that is,  $\mathbf{y} ={\cal N}\left( |\mathbf{F}\mathbf{x}|^2\right)$.
Here  $\cal N$  contains  photon noise and quantization, the latter using 8 bits per pixel. 
Note that even if $\mathbf{x}$ is a real-valued image, the measurement $\mathbf{y}$ has no phase.
This task is  ill-posed. 
The Hybrid Input-Output (HIO)~\cite{fienup1982phase} algorithm is frequently used in phase retrieval.
Let $\boldsymbol{\psi}$ be a random Gaussian vector.
HIO  initializes hundreds of possible candidate solutions $\{\mathbf{x}_l\}$ by $\mathbf{F}\mathbf{x}_l = \sqrt{\mathbf{y}}\exp(j2\pi \boldsymbol{\psi}_l)$. 
Then, HIO iteratively uses negative
feedback in Fourier space, in order to progressively force the candidate solutions to conform to the problem constraints.
The  candidate that holds  best  these constraints is the evaluated HIO solution.
\begin{figure}[t]
    \centering
    \includegraphics[width=\linewidth]{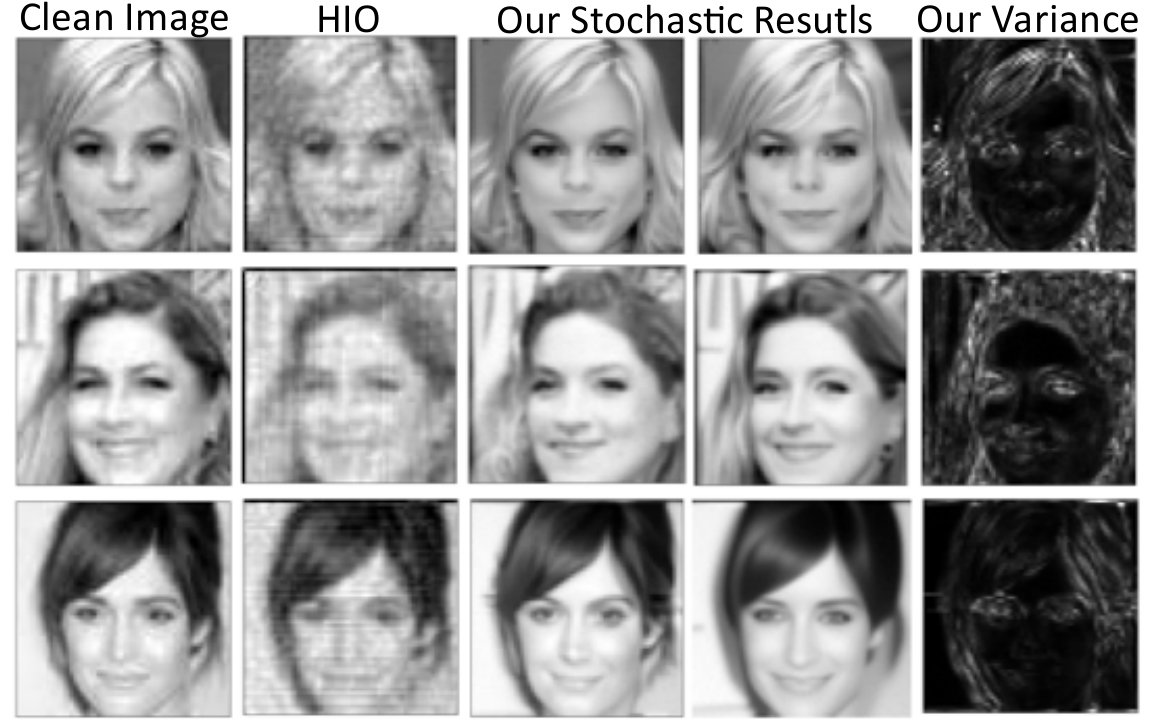}
    \caption{Phase retrieval results on CelebA images. A variance image is obtained by using 50 stochastic solutions.
    }
    \label{fig:my_label}
\end{figure}
\begin{figure}[t]
    \centering
    \includegraphics[width=1\linewidth]{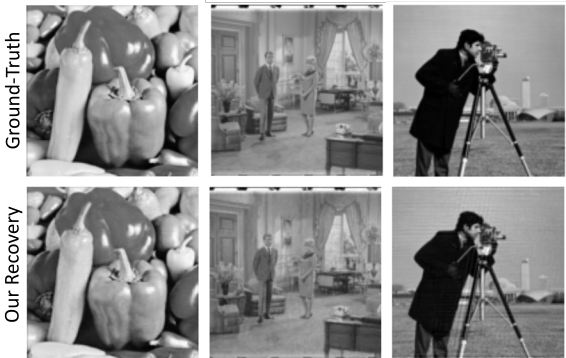}
    \caption{Phase retrieval qualitative results  for the dataset suggested in~\cite{metzler2018prdeep} \mbox{$(\sigma_0=4/255)$}. }
        \label{fig:pr_natural_dataset}
\end{figure}
\begin{table}[t]
\centering
\resizebox{.8\linewidth}{!}{
\centering
    \begin{tabular}{ l ccc } 
        \hline
         $\sigma_0=$ & {$2/255$}& {$3/255$}& {$4/255$} \\ 
        \hline
         HIO~\cite{fienup1982phase} & 22.2&19.9 & 17.9\\ 
         BM3D-prGAMP~\cite{metzler2016bm3d} &24.5 &23.2 & 21.0\\ 
         BM3D-ADMM~\cite{metzler2018prdeep}  &27.5 &24.5 & 21.8\\ 
         DnCNN-ADMM~\cite{metzler2018prdeep}  &29.3 &26.0 & 22.0\\ 
         prDeep~\cite{metzler2018prdeep}  & 28.4&28.5 & 26.4\\ 
          \bf{Ours}  &\bf{33.1} & \bf{32.6} & \bf{28.9}\\ 
        \hline
    \end{tabular}} 
   \caption{Phase retrieval PSNR results   for the ``natural" images dataset  in~\cite{metzler2018prdeep}, using different noise levels $\sigma_0$.}
    \label{tab:PSNRTablesPR}   
\end{table}
\label{sec:results_pr}

We compare our method (\cref{alg:two}) for  phase retrieval, vs. prior art.
Here, we initialize our algorithm with an extremely noisy version of the HIO solution. We found that the noisy initialization gives us better robustness during iterations, but still stochastically converges to a variety of  possible solutions (see \cref{fig:my_label}).
Additionally, we follow the experiment suggested in~\cite{metzler2018prdeep}. 
First, we trained $h_\Theta^{\rm Complex}(\mathbf{{o}}'|\sigma)$  with overlapping patches drawn from 400 images in the Berkeley Segmentation Dataset~\cite{martin2001database}. {Then, we evaluate  our method on a test set of six ``natural" images\cite{metzler2018prdeep}, three images are presented in \cref{fig:pr_natural_dataset}.
\cref{tab:PSNRTablesPR} compares the results.
In the supplementary material, we present more results  
and further details.}
\subsection{Fourier Ptychography: Real Experiment}
\label{sec:res_fp}
\begin{figure}[t]
\centering
\includegraphics[width=1\linewidth]{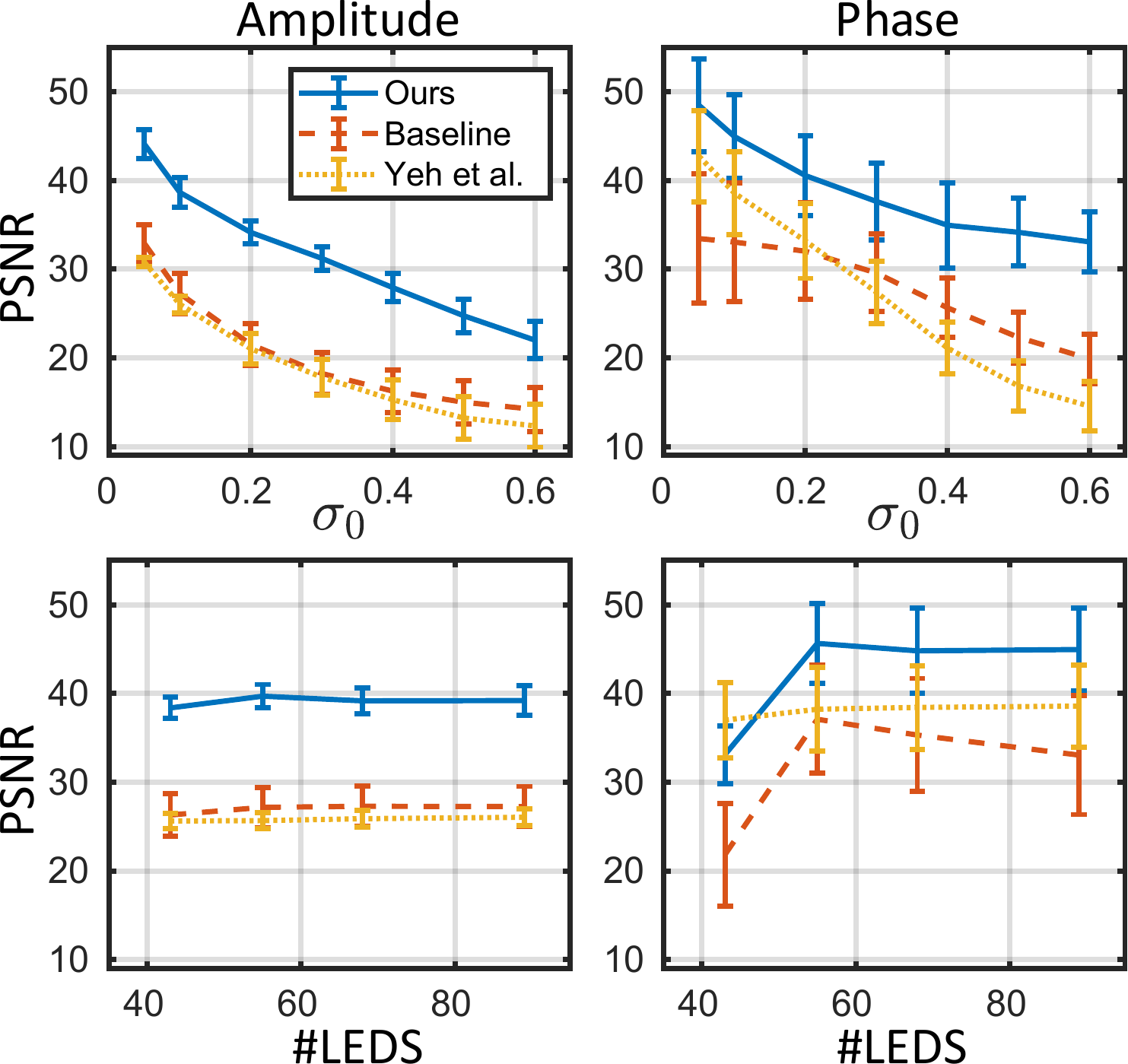}
  \caption{Fourier ptychography. Average PSNR for 10 complex-valued objects: [Top] Varying Poisson noise levels $\sigma_0$ and fixed $M=89$. [Bottom] 
  Varying  number of LEDs $M$  and a fixed noise level $\sigma_0=0.1$.
  [Baseline] \cref{alg:two} excluding the prior term $h_{\hat \Theta}^{\rm complex}$. Yeh et al.~\cite{yeh2015experimental} uses  quasi-Newton method.
  The bars stand for PSNR standard deviations.
  } 
  \label{fig:FP_complex_plots}
\end{figure}

\begin{figure}[t]
\centering
\includegraphics[width=.99\linewidth]{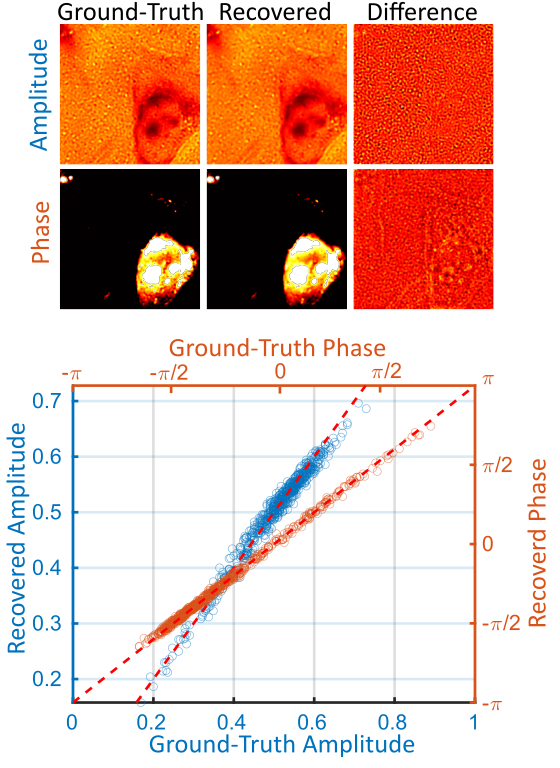}
  \caption{Fourier ptychography. Our results of a complex-valued restoration, for $M=89$ LEDs, $\sigma_0=0.1$ and 8 bit quantization. 
[Top] The amplitude and phase images' dynamic range $\in [0,1]$ and $[-\pi, \pi]$ respectively.
  [Bottom] Scatter plots for the amplitude and phase values, the dashed red lines represent a perfect recovery.} 
  \label{fig:scatter_complex}
\end{figure}

We further demonstrate our approach on the Fourier ptychography problem, described in \cref{sec:nonlinear_imaging}. 
We use data of real-world publicly available complex-valued biological samples~\cite{laura_data}.
Using a Fourier ptychography imaging system, 
the amplitude and phase of a sample of human bone cells are recovered~\cite{tian2014multiplexed}.
We randomly  divided  the recovered sample  into non-spatially overlapping patches, each of size $256\times 256$ pixels. 
Then, the patches were split randomly into training and test sets.
We train our denoiser over the training set, according to \cref{eq:denoiser_complex}.
For each  patch $\mathbf{o}$ in the test set, we simulate $M$ noisy measured images, each corresponding to a single LED in the optical system, according to \cref{eq:non_linear_transorm,eq:ptychography}.
\cref{fig:scatter_complex,fig:FP_complex_plots} show the results using \cref{alg:two}.
Further details about this test and more results are in the supplementary material.

\section{Discussion}
\label{sec:discuss}
We generalize annealed Langevin Dynamics to realistic optical imaging, which involves complex-valued objects and Poisson noise.
Our method   is generic and can be applied to a variety of  imaging transformations $\mathbf{H}$.
Furthermore, the pre-trained DNN in \cref{eq:denoiser_complex} is independent of  specific imaging system parameters $\sigma_0$ and $\mathbf{H}$ of \cref{eq:fm_non_linear}. Hence, the same DNN can be used for  various   optical  systems.

Dealing with  non-Gaussian noises, such as quantization noise or dead-time noise in SPAD-based imager,  is essential in real-world imaging. 
Hence, noise statistics create new and interesting challenges, specifically in low-light conditions.
There are more challenging forward models in imaging. These include imaging by coherent illumination (with speckles), partial-coherence, and  models that are nonlinear with respect to the illumination (e.g. two-photon microscopy). Other computer vision problems are highly nonlinear in the unknowns, such as tomography scatter fields~\cite{ronen2022variable}. 
Therefore, motivated by this paper, we hope that more research expands annealed Langevin Dynamics to additional challenging models  in real-world imaging. 


{\small
\bibliographystyle{ieee_fullname}
\bibliography{main}

\begin{thebibliography}{10}\itemsep=-1pt

\bibitem{akiyama2019first}
Kazunori Akiyama, Antxon Alberdi, Walter Alef, Keiichi Asada, Rebecca Azulay,
  Anne-Kathrin Baczko, David Ball, Mislav Balokovi{\'c}, John Barrett, Dan
  Bintley, et~al.
\newblock First {M}87 {E}vent {H}orizon {T}elescope results. {IV}. imaging the
  central supermassive black hole.
\newblock {\em The Astrophysical Journal Letters}, 875(1):L4, 2019.

\bibitem{attal2022towards}
Benjamin Attal and Matthew O'Toole.
\newblock Towards mixed-state coded diffraction imaging.
\newblock {\em IEEE Transactions on Pattern Analysis and Machine Intelligence},
  2022.

\bibitem{azzari2016variance}
Lucio Azzari and Alessandro Foi.
\newblock Variance stabilization for noisy+ estimate combination in iterative
  poisson denoising.
\newblock {\em IEEE signal processing letters}, 23(8):1086--1090, 2016.

\bibitem{blau2018perception}
Yochai Blau and Tomer Michaeli.
\newblock The perception-distortion tradeoff.
\newblock In {\em Proceedings of the IEEE conference on computer vision and
  pattern recognition}, pages 6228--6237, 2018.

\bibitem{bowman2012introduction}
Frank Bowman.
\newblock {\em Introduction to {B}essel {F}unctions}.
\newblock Courier Corporation, 2012.

\bibitem{buades2005non}
Antoni Buades, Bartomeu Coll, and J-M Morel.
\newblock A non-local algorithm for image denoising.
\newblock In {\em 2005 IEEE computer society conference on computer vision and
  pattern recognition (CVPR'05)}, volume~2, pages 60--65. Ieee, 2005.

\bibitem{byun2021fbi}
Jaeseok Byun, Sungmin Cha, and Taesup Moon.
\newblock {FBI}-denoiser: Fast blind image denoiser for poisson-gaussian noise.
\newblock In {\em Proceedings of the IEEE/CVF Conference on Computer Vision and
  Pattern Recognition}, pages 5768--5777, 2021.

\bibitem{cha2021deepphasecut}
Eunju Cha, Chanseok Lee, Mooseok Jang, and Jong~Chul Ye.
\newblock Deepphasecut: Deep relaxation in phase for unsupervised fourier phase
  retrieval.
\newblock {\em IEEE Transactions on Pattern Analysis and Machine Intelligence},
  44(12):9931--9943, 2021.

\bibitem{chan2022holocurtains}
Dorian Chan, Srinivasa~G Narasimhan, and Matthew O'Toole.
\newblock Holocurtains: Programming light curtains via binary holography.
\newblock In {\em Proceedings of the IEEE/CVF Conference on Computer Vision and
  Pattern Recognition (CVPR)}, pages 17886--17895, 2022.

\bibitem{chen2016trainable}
Yunjin Chen and Thomas Pock.
\newblock Trainable nonlinear reaction diffusion: A flexible framework for fast
  and effective image restoration.
\newblock {\em IEEE transactions on pattern analysis and machine intelligence},
  39(6):1256--1272, 2016.

\bibitem{choi2021ilvr}
Jooyoung Choi, Sungwon Kim, Yonghyun Jeong, Youngjune Gwon, and Sungroh Yoon.
\newblock Ilvr: Conditioning method for denoising diffusion probabilistic
  models.
\newblock {\em arXiv preprint arXiv:2108.02938}, 2021.

\bibitem{chung2022diffusion}
Hyungjin Chung, Jeongsol Kim, Michael~T Mccann, Marc~L Klasky, and Jong~Chul
  Ye.
\newblock Diffusion posterior sampling for general noisy inverse problems.
\newblock {\em arXiv preprint arXiv:2209.14687}, 2022.

\bibitem{chung2022improving}
Hyungjin Chung, Byeongsu Sim, Dohoon Ryu, and Jong~Chul Ye.
\newblock Improving diffusion models for inverse problems using manifold
  constraints.
\newblock {\em Advances in Neural Information Processing Systems},
  35:25683--25696, 2022.

\bibitem{chung2022come}
Hyungjin Chung, Byeongsu Sim, and Jong~Chul Ye.
\newblock Come-closer-diffuse-faster: Accelerating conditional diffusion models
  for inverse problems through stochastic contraction.
\newblock In {\em Proceedings of the IEEE/CVF Conference on Computer Vision and
  Pattern Recognition}, pages 12413--12422, 2022.

\bibitem{dremeau2015reference}
Ang{\'e}lique Dr{\'e}meau, Antoine Liutkus, David Martina, Ori Katz, Christophe
  Sch{\"u}lke, Florent Krzakala, Sylvain Gigan, and Laurent Daudet.
\newblock Reference-less measurement of the transmission matrix of a highly
  scattering material using a dmd and phase retrieval techniques.
\newblock {\em Optics Express}, 23(9):11898--11911, 2015.

\bibitem{eckert2018efficient}
Regina Eckert, Zachary~F Phillips, and Laura Waller.
\newblock Efficient illumination angle self-calibration in fourier
  ptychography.
\newblock {\em Applied Optics}, 57(19):5434--5442, 2018.

\bibitem{elad2006image}
Michael Elad and Michal Aharon.
\newblock Image denoising via sparse and redundant representations over learned
  dictionaries.
\newblock {\em IEEE Transactions on Image processing}, 15(12):3736--3745, 2006.

\bibitem{fienup1982phase}
James~R Fienup.
\newblock Phase retrieval algorithms: A comparison.
\newblock {\em Applied Optics}, 21(15):2758--2769, 1982.

\bibitem{funke2018large}
Jan Funke, Fabian Tschopp, William Grisaitis, Arlo Sheridan, Chandan Singh,
  Stephan Saalfeld, and Srinivas~C Turaga.
\newblock Large scale image segmentation with structured loss based deep
  learning for connectome reconstruction.
\newblock {\em IEEE transactions on pattern analysis and machine intelligence},
  41(7):1669--1680, 2018.

\bibitem{gu2014weighted}
Shuhang Gu, Lei Zhang, Wangmeng Zuo, and Xiangchu Feng.
\newblock Weighted nuclear norm minimization with application to image
  denoising.
\newblock In {\em Proceedings of the IEEE conference on computer vision and
  pattern recognition}, pages 2862--2869, 2014.

\bibitem{huang2021neighbor2neighbor}
Tao Huang, Songjiang Li, Xu Jia, Huchuan Lu, and Jianzhuang Liu.
\newblock Neighbor2neighbor: Self-supervised denoising from single noisy
  images.
\newblock In {\em Proceedings of the IEEE/CVF conference on Computer Vision and
  Pattern Recognition (CVPR)}, pages 14781--14790, 2021.

\bibitem{hyder2020solving}
Rakib Hyder, Zikui Cai, and M~Salman Asif.
\newblock Solving phase retrieval with a learned reference.
\newblock In {\em European Conference on Computer Vision (ECCV)}, pages
  425--441. Springer, 2020.

\bibitem{kadkhodaie2021stochastic}
Zahra Kadkhodaie and Eero Simoncelli.
\newblock Stochastic solutions for linear inverse problems using the prior
  implicit in a denoiser.
\newblock {\em Advances in Neural Information Processing Systems (NIPS)},
  34:13242--13254, 2021.

\bibitem{kawar2022denoising}
Bahjat Kawar, Michael Elad, Stefano Ermon, and Jiaming Song.
\newblock Denoising diffusion restoration models.
\newblock {\em Advances in Neural Information Processing Systems},
  35:23593--23606, 2022.

\bibitem{kawar2021snips}
Bahjat Kawar, Gregory Vaksman, and Michael Elad.
\newblock {SNIPS}: Solving noisy inverse problems stochastically.
\newblock {\em Advances in Neural Information Processing Systems (NIPS)},
  34:21757--21769, 2021.

\bibitem{kawar2021stochastic}
Bahjat Kawar, Gregory Vaksman, and Michael Elad.
\newblock Stochastic image denoising by sampling from the posterior
  distribution.
\newblock In {\em Proceedings of the IEEE/CVF International Conference on
  Computer Vision (ICCV)}, pages 1866--1875, 2021.

\bibitem{khademi2021self}
Wesley Khademi, Sonia Rao, Clare Minnerath, Guy Hagen, and Jonathan Ventura.
\newblock Self-supervised poisson-gaussian denoising.
\newblock In {\em Proceedings of the IEEE/CVF Winter Conference on Applications
  of Computer Vision}, pages 2131--2139, 2021.

\bibitem{krull2019noise2void}
Alexander Krull, Tim-Oliver Buchholz, and Florian Jug.
\newblock Noise2void-learning denoising from single noisy images.
\newblock In {\em Proceedings of the IEEE/CVF conference on computer vision and
  pattern recognition}, pages 2129--2137, 2019.

\bibitem{kumar2019low}
Prashanth~G Kumar and Rajiv Ranjan~Sahay.
\newblock Low rank poisson denoising ({LRPD}): A low rank approach using split
  bregman algorithm for poisson noise removal from images.
\newblock In {\em Proceedings of the IEEE/CVF Conference on Computer Vision and
  Pattern Recognition Workshops (WCVPR)}, pages 0--0, 2019.

\bibitem{liu2018non}
Ding Liu, Bihan Wen, Yuchen Fan, Chen~Change Loy, and Thomas~S Huang.
\newblock Non-local recurrent network for image restoration.
\newblock {\em Advances in neural information processing systems}, 31, 2018.

\bibitem{martin2001database}
David Martin, Charless Fowlkes, Doron Tal, and Jitendra Malik.
\newblock A database of human segmented natural images and its application to
  evaluating segmentation algorithms and measuring ecological statistics.
\newblock In {\em Proceedings Eighth IEEE International Conference on Computer
  Vision (ICCV)}, volume~2, pages 416--423. IEEE, 2001.

\bibitem{metzler2018prdeep}
Christopher Metzler, Phillip Schniter, Ashok Veeraraghavan, and Richard
  Baraniuk.
\newblock Prdeep: Robust phase retrieval with a flexible deep network.
\newblock In {\em International Conference on Machine Learning}, pages
  3501--3510. PMLR, 2018.

\bibitem{metzler2016bm3d}
Christopher~A Metzler, Arian Maleki, and Richard~G Baraniuk.
\newblock {BM3D-PRGAMP}: Compressive phase retrieval based on {BM3D} denoising.
\newblock In {\em 2016 IEEE International Conference on Image Processing
  (ICIP)}, pages 2504--2508. IEEE, 2016.

\bibitem{mildenhall2018burst}
Ben Mildenhall, Jonathan~T Barron, Jiawen Chen, Dillon Sharlet, Ren Ng, and
  Robert Carroll.
\newblock Burst denoising with kernel prediction networks.
\newblock In {\em Proceedings of the IEEE conference on Computer Vision and
  Pattern Recognition}, pages 2502--2510, 2018.

\bibitem{moseley2021extreme}
Ben Moseley, Valentin Bickel, Ignacio~G L{\'o}pez-Francos, and Loveneesh Rana.
\newblock Extreme low-light environment-driven image denoising over permanently
  shadowed lunar regions with a physical noise model.
\newblock In {\em Proceedings of the IEEE/CVF Conference on Computer Vision and
  Pattern Recognition}, pages 6317--6327, 2021.

\bibitem{ratner2007illumination}
Netanel Ratner and Yoav~Y Schechner.
\newblock Illumination multiplexing within fundamental limits.
\newblock In {\em IEEE Conference on Computer Vision and Pattern Recognition
  (CVPR)}, pages 1--8. IEEE, 2007.

\bibitem{rodenburg2019ptychography}
John Rodenburg and Andrew Maiden.
\newblock Ptychography.
\newblock {\em Springer Handbook of Microscopy}, pages 819--904, 2019.

\bibitem{romano2017little}
Yaniv Romano, Michael Elad, and Peyman Milanfar.
\newblock The little engine that could: Regularization by denoising ({RED}).
\newblock {\em SIAM Journal on Imaging Sciences}, 10(4):1804--1844, 2017.

\bibitem{ronen2022variable}
Roi Ronen, Vadim Holodovsky, and Yoav~Y Schechner.
\newblock Variable imaging projection cloud scattering tomography.
\newblock {\em IEEE Transactions on Pattern Analysis and Machine Intelligence},
  2022.

\bibitem{saha2022lwgnet}
Atreyee Saha, Salman~S Khan, Sagar Sehrawat, Sanjana~S Prabhu, Shanti
  Bhattacharya, and Kaushik Mitra.
\newblock Lwgnet-learned wirtinger gradients for fourier ptychographic phase
  retrieval.
\newblock In {\em European Conference on Computer Vision (ECCV)}, pages
  522--537. Springer, 2022.

\bibitem{shamshad2019adaptive}
Fahad Shamshad, Asif Hanif, Farwa Abbas, Muhammad Awais, and Ali Ahmed.
\newblock Adaptive ptych: Leveraging image adaptive generative priors for
  subsampled fourier ptychography.
\newblock In {\em Proceedings of the IEEE/CVF International Conference on
  Computer Vision Workshops (WICCV)}, pages 0--0, 2019.

\bibitem{shechtman2015phase}
Yoav Shechtman, Yonina~C Eldar, Oren Cohen, Henry~Nicholas Chapman, Jianwei
  Miao, and Mordechai Segev.
\newblock Phase retrieval with application to optical imaging: {A} contemporary
  overview.
\newblock {\em IEEE Signal Processing Magazine}, 32(3):87--109, 2015.

\bibitem{song2019generative}
Yang Song and Stefano Ermon.
\newblock Generative modeling by estimating gradients of the data distribution.
\newblock {\em Advances in Neural Information Processing Systems (NIPS)}, 32,
  2019.

\bibitem{song2020improved}
Yang Song and Stefano~and Ermon.
\newblock Improved techniques for training score-based generative models.
\newblock {\em Advances in Neural Information Processing Systems (NIPS},
  33:12438--12448, 2020.

\bibitem{song2021solving}
Yang Song, Liyue Shen, Lei Xing, and Stefano Ermon.
\newblock Solving inverse problems in medical imaging with score-based
  generative models.
\newblock {\em arXiv preprint arXiv:2111.08005}, 2021.

\bibitem{laura_data}
Lei Tian, Xiao Li, Kannan Ramchandran, and Laura Waller.
\newblock {\em (Dataset) LED array 2D Fourier Ptychography. Available online}.
\newblock \url{http://gigapan.com/gigapans/170956}.

\bibitem{tian2014multiplexed}
Lei Tian, Xiao Li, Kannan Ramchandran, and Laura Waller.
\newblock Multiplexed coded illumination for {F}ourier ptychography with an
  {LED} array microscope.
\newblock {\em Biomedical Optics Express}, 5(7):2376--2389, 2014.

\bibitem{wang2009mean}
Zhou Wang and Alan~C Bovik.
\newblock Mean squared error: Love it or leave it? a new look at signal
  fidelity measures.
\newblock {\em IEEE Signal Processing Magazine}, 26(1):98--117, 2009.

\bibitem{welling2011bayesian}
Max Welling and Yee~W Teh.
\newblock Bayesian learning via stochastic gradient langevin dynamics.
\newblock In {\em Proceedings of the International Conference on Machine
  Learning (ICML)}, pages 681--688, 2011.

\bibitem{wu2019wish}
Yicheng Wu, Manoj~Kumar Sharma, and Ashok Veeraraghavan.
\newblock Wish: wavefront imaging sensor with high resolution.
\newblock {\em Light: Science \& Applications}, 8(1):44, 2019.

\bibitem{xue2022convergence}
Duoduo Xue, Ziyang Zheng, Wenrui Dai, Chenglin Li, Junni Zou, and Hongkai
  Xiong.
\newblock On the convergence of non-convex phase retrieval with denoising
  priors.
\newblock {\em IEEE Transactions on Signal Processing}, 70:4424--4439, 2022.

\bibitem{yeh2015experimental}
Li-Hao Yeh, Jonathan Dong, Jingshan Zhong, Lei Tian, Michael Chen, Gongguo
  Tang, Mahdi Soltanolkotabi, and Laura Waller.
\newblock Experimental robustness of fourier ptychography phase retrieval
  algorithms.
\newblock {\em Optics express}, 23(26):33214--33240, 2015.

\bibitem{zhang2021physics}
Feilong Zhang, Xianming Liu, Cheng Guo, Shiyi Lin, Junjun Jiang, and Xiangyang
  Ji.
\newblock Physics-based iterative projection complex neural network for phase
  retrieval in lensless microscopy imaging.
\newblock In {\em Proceedings of the IEEE/CVF Conference on Computer Vision and
  Pattern Recognition (CVPR)}, pages 10523--10531, 2021.

\bibitem{zhang2017beyond}
Kai Zhang, Wangmeng Zuo, Yunjin Chen, Deyu Meng, and Lei Zhang.
\newblock Beyond a gaussian denoiser: Residual learning of deep cnn for image
  denoising.
\newblock {\em IEEE transactions on image processing}, 26(7):3142--3155, 2017.

\bibitem{zhang2019poisson}
Yide Zhang, Yinhao Zhu, Evan Nichols, Qingfei Wang, Siyuan Zhang, Cody Smith,
  and Scott Howard.
\newblock A {P}oisson-{G}aussian denoising dataset with real fluorescence
  microscopy images.
\newblock In {\em Proceedings of the IEEE/CVF Conference on Computer Vision and
  Pattern Recognition}, pages 11710--11718, 2019.

\bibitem{zheng2013wide}
Guoan Zheng, Roarke Horstmeyer, and Changhuei Yang.
\newblock Wide-field, high-resolution fourier ptychographic microscopy.
\newblock {\em Nature Photonics}, 7(9):739--745, 2013.

\bibitem{zoran2011learning}
Daniel Zoran and Yair Weiss.
\newblock From learning models of natural image patches to whole image
  restoration.
\newblock In {\em 2011 international conference on computer vision}, pages
  479--486. IEEE, 2011.

\end{thebibliography}
}

\end{document}